\documentclass{article}
\usepackage[a4paper,margin=3cm]{geometry}  
\usepackage{amsfonts}
\usepackage{graphicx}
\usepackage{amsmath}
\usepackage{makecell}
\usepackage{float}
\usepackage{authblk}
\usepackage{tikz}
\usetikzlibrary{matrix}
\usetikzlibrary{positioning,arrows.meta,fit,shapes.multipart}

\title{Interpretable Computer Vision for Defect Detection in X-ray Tomography of Aerospace SiC/SiC Composites}
\author[1]{Antonio Pe\~na Corredor}
\author[1]{Julien Lesseur}
\author[2]{Romain Nunez}
\author[3]{Paul Rivalland}
\author[2]{Thomas Philippe}

\affil[1]{IRT Saint Exup\'ery, Toulouse, France}
\affil[2]{Safran Ceramics, Le Haillan, France}
\affil[3]{Safran Engineering Services, Le Haillan, France}

\begin{document}

\maketitle

\begin{abstract}
X-ray computed tomography (XCT) is widely used for non-destructive testing (NDT) of aerospace SiC/SiC composites, but current inspection workflows rely heavily on expert visual assessment and provide limited traceability for accept/reject decisions. Deep convolutional networks can detect defects automatically, yet are typically deployed as black boxes that output a defect probability without a structured explanation.

In this work, we develop an interpretable prototype-based framework for defect detection in XCT of SiC/SiC parts. We first construct and release a curated patch dataset from four defect-rich laboratory specimens and train a family of residual networks (ResNet-18 to ResNet-152) under a common protocol. All backbones achieve near-ceiling performance on the held-out test set (ROC--AUC $\geq 0.991$, PR--AUC $\geq 0.986$), with ResNet-50 offering the best trade-off between accuracy, F$_1$-score, specificity and calibration, and therefore serves as the baseline.

We then introduce a prototype-based extension (p--ResNet-50) in which six learned prototypes are explicitly aligned with expert-defined semantic categories (pure air, healthy matrix, matrix with air interface, pores, line-like defects, mixed pores and lines). Its training objective includes novel anchor- and medoid-based regularization terms that tether prototypes to expert-selected examples and mitigate prototype collapse. The prototype network attains performance comparable to the black-box baseline (accuracy $0.957$ vs.\ $0.959$, F$_1$ $0.939$ vs.\ $0.945$, ROC--AUC $0.994$ vs.\ $0.993$) while providing case-based explanations via representative training patches and patch-wise prototype attributions, with a modest accuracy--interpretability trade-off manifesting as slightly reduced sensitivity in exchange for higher precision and specificity. Uniform Manifold Approximation and Projection (UMAP) visualizations reveal that the latent space naturally organizes into two disconnected ``islands''---one grouping low-contrast non-defect patches, the other grouping high-contrast patches where defect and non-defect categories coexist---within which prototype training carves compact, semantically coherent sub-domains and localizes the zones of uncertainty where misclassifications are concentrated, delineating the model's domain of applicability.

The resulting framework yields not only accurate defect maps, but also domain-consistent, interpretable predictions that can be integrated into NDT decision-making workflows, making component sanctioning and acceptance more transparent and auditable.

\end{abstract}

\section{Introduction}

X-ray computed tomography (XCT) has become a key modality for non-destructive testing (NDT) of industrial components, enabling the detection and characterization of internal defects such as pores, cracks and inclusions~\cite{duPlessis2016_comparison,kastner2015_xct_ndt}. In ceramic-matrix composites based on silicon carbide (SiC/SiC), which are increasingly considered for aero-engine and high-temperature aerospace applications~\cite{hu2024_sicf_pip}, the situation is particularly challenging: engineered fiber architectures, interphases and controlled porosity give rise to highly heterogeneous microstructures, within which manufacturing defects and in-service damage interact with the reinforcement architecture~\cite{thornton2019_sicsic_microct,wan2019_ccsic_xct}. For safety-critical applications, decisions to accept or reject a part often rely on detailed inspection of tomographic volumes by expert operators, who visually scan cross-sections and mentally integrate information over the full 3D component. This workflow is time-consuming, difficult to scale, and susceptible to inter- and intra-operator variability and other human-factor limitations~\cite{see2012_visualInspection,bertovic2016_humanFactorsNDT}, especially when defects are small, numerous, or embedded in complex composite microstructures.

Automated analysis methods that can reliably detect and quantify defects therefore have the potential to substantially improve throughput and reproducibility. In particular, convolutional neural network (CNN)–based segmentation and classification pipelines have been shown to accelerate XCT-based inspection of additively manufactured parts while maintaining or improving sensitivity to pores and lack-of-fusion defects~\cite{wong2022_unetAM,ziabari2023_rapidXCTAM,xu2024_highperfSeg}, and similar gains have been reported in other NDT modalities, where machine-learning models reduce manual interpretation time and operator-to-operator variability~\cite{klewe2024_gpr_ml}.

Within this landscape, deep convolutional neural networks have emerged as the dominant approach for image-based inspection and NDT, including XCT defect detection and segmentation~\cite{wong2022_unetAM,ziabari2023_rapidXCTAM}. Residual networks (ResNets), in particular, offer a powerful and flexible backbone for learning discriminative representations from image patches and aggregating these predictions into defect maps at the part level~\cite{he2016_resnet}. Yet in their standard form, these models are typically used as black boxes: they output a probability of ``defect'' versus ``non-defect'', but offer limited insight into why a given region is classified as defective or under which conditions their predictions can be trusted.

Post-hoc explanation tools such as gradient-based saliency maps~\cite{simonyan2014_saliency,zeiler2014_visualizing} or Grad-CAM-style localization methods~\cite{selvaraju2017_gradcam} can highlight image regions that contribute to a prediction, but they do not decompose decisions into explicit, domain-meaningful defect types nor provide directly actionable, quantitative summaries for sanctioning parts~\cite{rudin2019_stopExplaining,murdoch2019_definitions}.

For industrial NDT, this lack of structured interpretability is a major limitation: experts must be able to understand and justify why a given region is flagged as defective in order to trust and operationalize automated decisions~\cite{rudin2019_stopExplaining,ukwaththa2024_xai_xct}. In practice, inspectors are not only interested in whether a part is defective, but also in the nature of the defects (e.g., pores vs.\ line-like features), their spatial distribution within the component, and their severity relative to acceptance criteria and service requirements~\cite{duPlessis2016_comparison}. At the same time, regulatory and certification contexts increasingly demand auditable models whose domain of applicability can be characterized and whose failure modes can be systematically analyzed~\cite{guidotti2018_surveyExpl,doshi2017_rigorous}.
These needs motivate approaches that combine the discriminative power of modern CNNs with explicit, case-based representations aligned with established defect taxonomies, enabling quantitative, decision-oriented summaries rather than purely opaque scores~\cite{chen2019_protopnet,ukwaththa2024_xai_xct}.

Prototype-based neural networks provide such a mechanism by augmenting a standard backbone with a set of learned prototypes in the latent space, each associated with a human-interpretable concept~\cite{snell2017_prototypical,chen2019_protopnet,nauta2021_prototree,rymarczyk2021_prototype}. During inference, predictions are obtained by comparing an input patch to these prototypes, so that decisions can be explained in terms of similarity to concrete training examples rather than abstract feature activations. In the context of XCT-based NDT, this opens the possibility of tying each prediction to semantic defect categories (e.g., pores, line-like defects, healthy matrix) and of analyzing the structure of the learned feature space in terms that are directly meaningful to domain experts. In our setting, domain knowledge about relevant defect morphologies is explicitly encoded in the definition and initialization of prototype types, following the broader principle of constraining data-driven models with domain knowledge that has proved effective in manufacturing applications~\cite{pena2025_physInformedLPBF}.

\definecolor{proto0}{HTML}{D62728} 
\definecolor{proto1}{HTML}{FFF200} 
\definecolor{proto2}{HTML}{FF7F0E} 
\definecolor{proto3}{HTML}{00D6AB} 
\definecolor{proto4}{HTML}{8E2BFF} 
\definecolor{proto5}{HTML}{B1B7FC} 
\definecolor{defectColor}{HTML}{00008B}   
\definecolor{nondefColor}{HTML}{FFA500}   

\begin{figure}[H]
    \centering
    \begin{tikzpicture}[
        scale=0.85,
        every node/.style={transform shape, align=center},
        font=\small,
        >={Stealth},
        slice/.style={inner sep=0pt},
        net/.style={draw, thick, rounded corners,
                    minimum width=2.6cm, minimum height=0.9cm,
                    fill=green!20},
        maps/.style={inner sep=0pt},
        binbox/.style={inner sep=0pt},
        prob/.style={draw, thick, minimum width=0.5cm, minimum height=0.5cm},
        bigproto/.style={draw, line width=2pt,
                         minimum width=1.5cm, minimum height=1.5cm,
                         fill=white, inner sep=0pt}
    ]

    \def\yt{4}
    \def\yb{-4}

    \node[slice,label=above:{XCT slice}] (xct) at (0,0)
        {\includegraphics[width=0.18\textwidth]{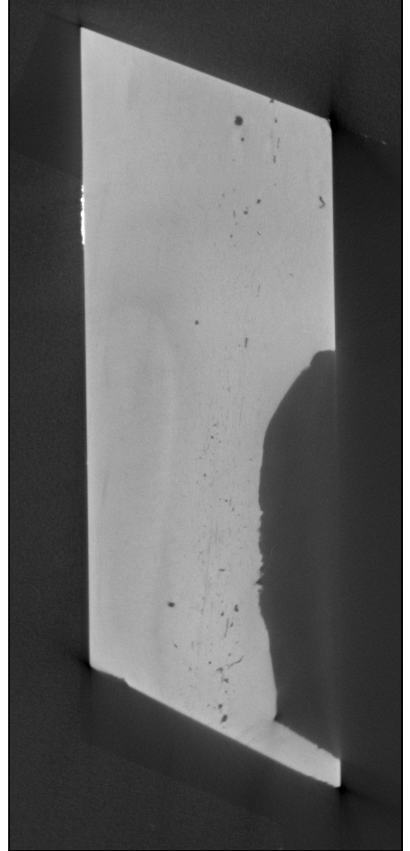}};

    \node[net] (resnet)  at (4,\yt)  {ResNet};
    \node[net] (presnet) at (4,\yb)  {p--ResNet};

    \node[prob,fill=white,
          label=above:{Defect\\probability}] (defprob) at (7,\yt) {};

    \node[prob,fill=proto0] (p0) at (7,\yb+1.25) {};
    \node[prob,fill=proto1] (p1) at (7,\yb+0.75) {};
    \node[prob,fill=proto2] (p2) at (7,\yb+0.25) {};
    \node[prob,fill=proto3] (p3) at (7,\yb-0.25) {};
    \node[prob,fill=proto4] (p4) at (7,\yb-0.75) {};
    \node[prob,fill=proto5] (p5) at (7,\yb-1.25) {};
    \node[above=0.15cm of p0] {Prototype\\probabilities};

    \node[maps,
          label=above:{Prototype-based\\prediction}] (maps_p)
          at (9.5,\yb)
          {\includegraphics[width=0.18\textwidth]{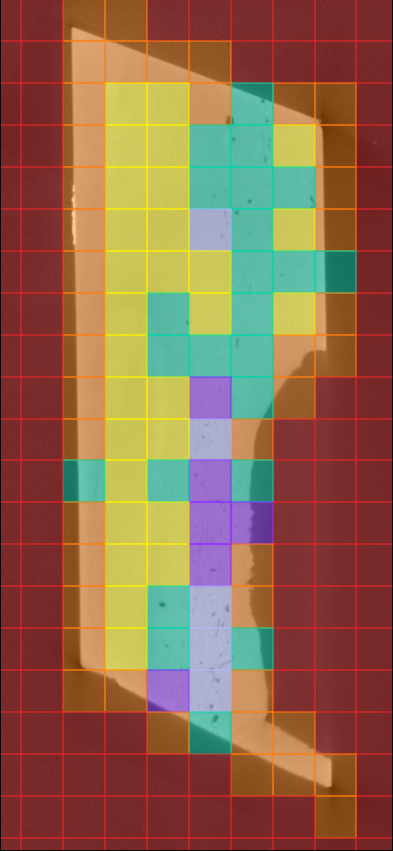}};

    \node[binbox,
          label=above:{Binary classification}] (bin_res)
          at (13,\yt)
          {\includegraphics[width=0.18\textwidth]{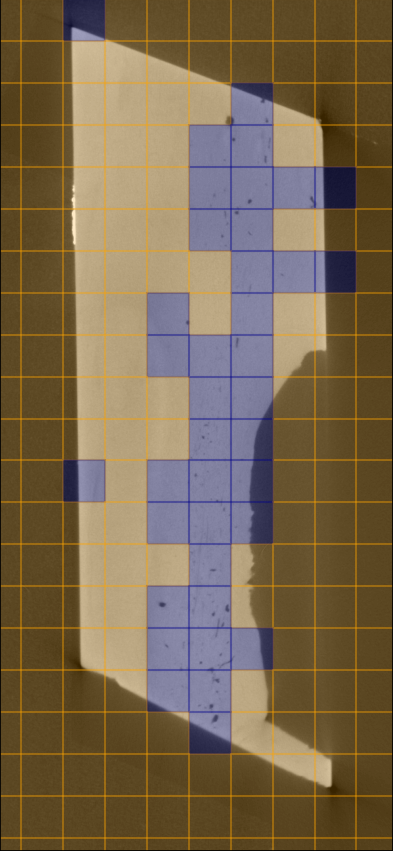}};

    \node[binbox] (bin_p) at (13,\yb)
          {\includegraphics[width=0.18\textwidth]{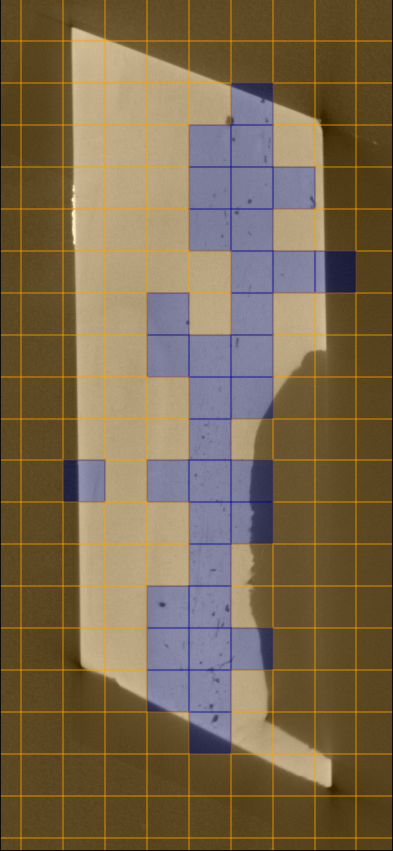}};

    \matrix (binlegend) at (13,0) [row sep=0.12cm, column sep=0.25cm] {
        \node[prob,draw=none,fill=defectColor] {}; & \node{defect}; \\
        \node[prob,draw=none,fill=nondefColor] {}; & \node{non-defect}; \\
    };

    \matrix (legend) at (7.5,-9.0) [row sep=0.15cm, column sep=0.4cm, nodes={anchor=base}] {
        \node[bigproto,draw=proto0, anchor=center]
            {\includegraphics[width=1.3cm]{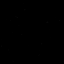}}; &
        \node[bigproto,draw=proto1, anchor=center]
            {\includegraphics[width=1.3cm]{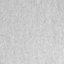}}; &
        \node[bigproto,draw=proto2, anchor=center]
            {\includegraphics[width=1.3cm]{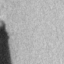}}; &
        \node[bigproto,draw=proto3, anchor=center]
            {\includegraphics[width=1.3cm]{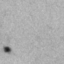}}; &
        \node[bigproto,draw=proto4, anchor=center]
            {\includegraphics[width=1.3cm]{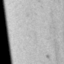}}; &
        \node[bigproto,draw=proto5, anchor=center]
            {\includegraphics[width=1.3cm]{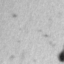}}; \\
        \node{air}; &
        \node{matrix}; &
        \node{air+matrix}; &
        \node{pores}; &
        \node{lines}; &
        \node{pores+lines}; \\
    };

    \node[left=0.2cm of legend.west, yshift=0.3cm] {Prototypes:};

    \draw[->,thick] (xct.east) -- ++(0.8,0) |- (resnet.west);
    \draw[->,thick] (xct.east) -- ++(0.8,0) |- (presnet.west);

    \draw[->,thick] (resnet.east) -- (defprob.west);
    \draw[->,thick] (defprob.east) --
        node[midway,above]{threshold ($t$)} (bin_res.west);

    \draw[->,thick] (presnet.east) -- (6.75, \yb); 
    
    \draw[->,thick] (7.25, \yb) -- (maps_p.west);
    \draw[->,thick] (maps_p.east) --
        node[midway,above]{$t$} (bin_p.west);

    \end{tikzpicture}
    \caption{Schematic overview of the defect–detection framework. The prototype-based p--ResNet produces attribution scores across six semantic classes, which are visually aligned and calibrated via thresholding.}
    \label{fig:overview}
\end{figure}
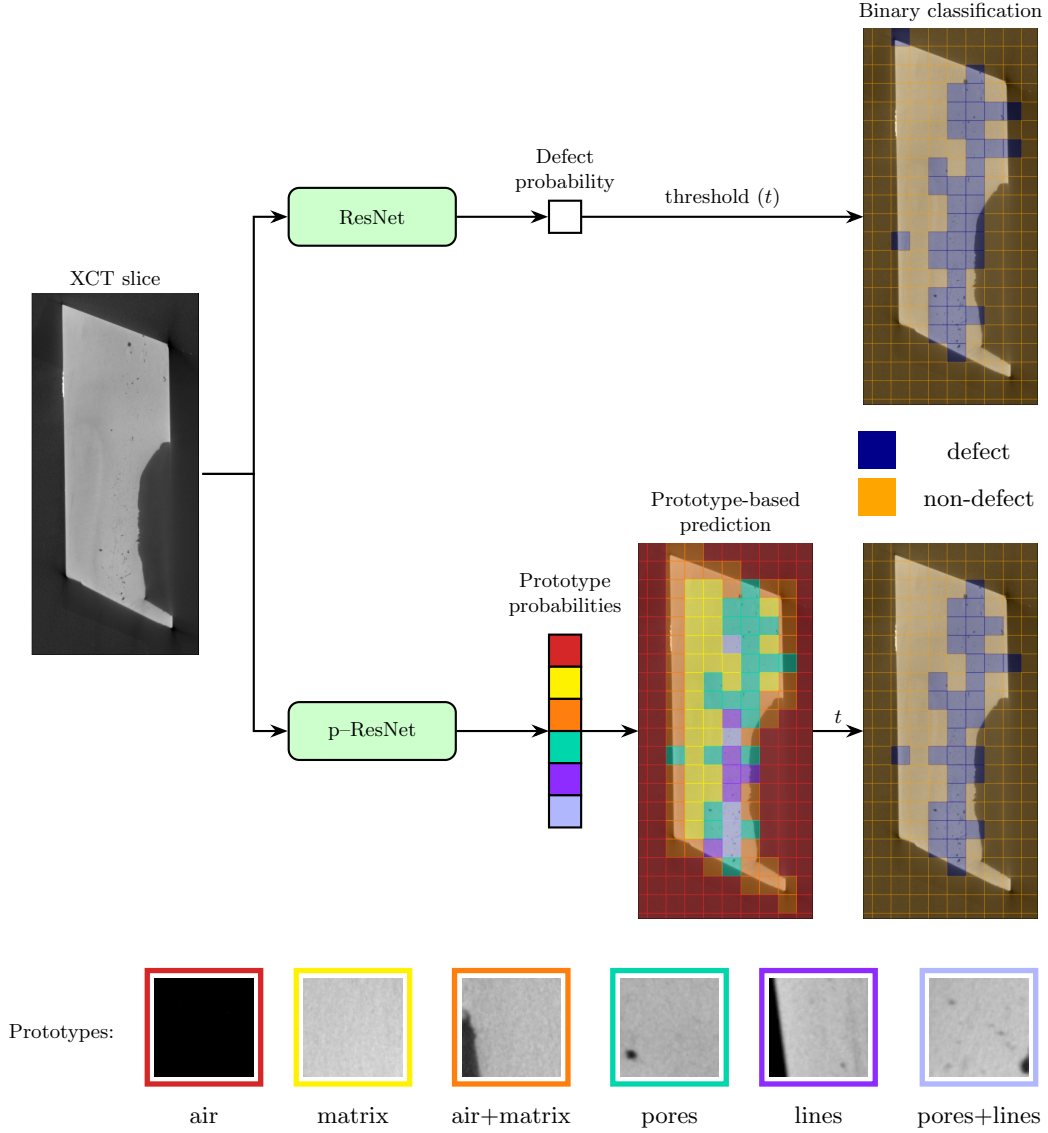

In this work, we investigate the use of prototype-based networks for explainable defect detection in XCT data, as schematized in Fig.~\ref{fig:overview}. We first establish a strong supervised baseline by training and calibrating a family of ResNet classifiers on patch-level defect labels, and systematically comparing their performance under a common evaluation protocol. We then select ResNet-50 as a backbone and construct a prototype-based variant (p--ResNet-50) in which learned prototypes are explicitly aligned with six expert-defined semantic categories (air, healthy matrix, pores, line-like defects, mixed pores and lines). Beyond standard metrics, we analyze calibration quality, inspect learned prototypes and their nearest training ``anchors'', and compare patch-wise predictions between the black-box and prototype-based models.

Our contributions are threefold.
(i) We construct a curated patch-level XCT dataset of defect-rich SiC/SiC specimens,
with standardized train/validation/test splits and expert-verified labels. (ii) We use this dataset
to perform a thorough quantitative assessment of residual backbones for patch-level
defect detection in XCT, including confidence intervals and calibration analysis;
within this setting, ResNet-50 emerges as a strong yet efficient baseline.
(iii) We design and evaluate a prototype-based extension of ResNet-50 whose training
objective incorporates an anchor-based regularization term and temperature control,
thereby preserving competitive discriminative performance while offering semantically
grounded, case-based explanations of individual predictions, as well as patch-wise
defect maps.
(iv) We analyze the latent spaces of both models using Uniform Manifold Approximation
and Projection (UMAP) projections~\cite{mcinnes2018_umap} colored by prototype
attribution, revealing compact, semantically coherent regions, localized
``uncertainty islands'', and clear strengths and weaknesses with respect to different
defect types. This embedding-level analysis provides a concrete way to audit the
model and to delineate its domain of applicability, thereby supporting more informed
use of deep learning in NDT workflows.

Together, these results suggest that prototype-based architectures constitute a promising compromise between accuracy and interpretability for industrial XCT inspection: they retain much of the discriminative power of black-box CNNs while yielding explanations and quantitative summaries that are directly usable by NDT experts when sanctioning or accepting parts.

\section{Materials and methods}

\subsection*{Data acquisition}

The study used SiC/SiC matrix composite parts manufactured
specifically for this work. Although produced in the laboratory, their
geometry, lay-up and apparent density were designed to resemble
aeronautical and aerospace-grade components routinely inspected by
X-ray tomography~\cite{hu2024_sicf_pip,wan2019_ccsic_xct}. In contrast
to production parts, they were intentionally made defect-rich to expose
the models to a broader variety of defect appearances and to obtain
informative test sets. Four different specimens were employed for model
training, validation and testing. Each reconstructed slice had an in-plane resolution
of $930 \times 1485$ pixels and was stored with 16-bit intensity encoding.

Two main defect morphologies were observed:
(i) pore-like defects, usually with strong intensity contrast relative
to the matrix; and
(ii) laminar or line-like defects, typically fainter and more elongated,
with less clear boundaries. For modelling, all defect morphologies were
merged into a single \emph{defect} class, yielding a binary task (defect
vs.\ non-defect). In a later stage, semantic subtypes (pores, line-like
defects, mixed patterns) are employed to define prototype types.

A semantic segmentation formulation was considered but discarded because
(i) experts showed poor agreement on the precise extent of laminar
defects, leading to unstable masks, and
(ii) pixel-level annotation was substantially more time-consuming than
patch-level labeling. The problem was therefore reformulated as a binary
classification on $64\times 64$ tomographic patches. To accelerate
experimentation and remain compatible with deployment on modest
hardware, 2D patch-based models were selected.

On the four available volumes, 51\,429 patches were sampled
pseudo-randomly with a bias towards intensity values near the median
matrix intensity. Among them, 12\,410 had a very low mean intensity
($<30/255$), well below the typical matrix range ($101$–$126/255$), and
were automatically labeled as non-defect (class~0). The remaining
39\,019 patches were inspected individually. Annotators could visualize
each patch in its volume context to disambiguate subtle cases.

Across the full raw dataset, approximately $2.7\%$ of patches contained
a defect. Direct training on this distribution ($\approx 36{:}1$
non-defect:defect) would lead to strong class imbalance. For model
development, we therefore sub-sampled non-defect patches to target
approximately a $2{:}1$ non-defect:defect ratio. The sub-sampling
probability was made proportional to the distance from the median matrix
intensity. After rebalancing and expert verification, the final curated
dataset comprised 1\,400 defect patches and 2\,784 non-defect patches
($80\%$ for training, $10\%$ for validation and $10\%$ for testing). The validation set was also employed for model calibration. \footnote{The authors
considered holding out a separate calibration split, but given the
limited dataset size this would have excessively reduced the amount of
training and validation data. We therefore followed the common practice
of reusing the validation set for both model selection and calibration.} 

\subsection*{Baseline CNN classifiers}

We first trained a family of residual convolutional neural networks
(ResNet-18, -34, -50, -101 and -152) on the patch-level binary
classification task. All networks were adapted to single-channel
(grayscale) inputs by modifying the first convolutional layer. Training
hyperparameters were kept identical across depths.

Each model was optimized with a binary cross-entropy loss
(\texttt{BCEWithLogitsLoss}) using the Adam optimizer (learning rate
$5\times 10^{-5}$, batch size $8$). A
\texttt{ReduceLROnPlateau} scheduler (factor $0.5$, patience $10$,
minimum learning rate $10^{-7}$) monitored the validation loss and
reduced the learning rate when it plateaued. Online data augmentation on
training patches consisted of random horizontal and vertical flips
($p=0.5$), small in-plane rotations (up to $15^\circ$, $p=0.25$), and
per-patch normalization to zero mean and unit variance. Validation and
test patches were only normalized.

For each model, the decision threshold on the predicted defect
probability was selected on the validation set by maximizing the
F\textsubscript{1} score. Probabilities were post-hoc calibrated on the
same validation split using temperature scaling ~\cite{guo2017_calibration}. As shown in the Results
section, the ResNet-50 model led to the best results and was therefore
chosen as backbone for the prototype network.

\subsection*{Prototype-based classifier (p--ResNet-50)}

We constructed a prototype-based model (p--ResNet-50) on top of the
ResNet-50 backbone to obtain interpretable predictions while preserving
detection performance. The model produces a binary decision (defect vs.\
non-defect) together with a distribution over a small set of learned,
human-readable prototypes.

The ResNet-50 encoder is used up to the global average pooling layer,
with a single-channel first convolution as in the baseline. The final
fully connected layer is removed, yielding for each $64\times 64$ patch
$x$ a 2048-dimensional embedding $z(x)\in\mathbb{R}^{2048}$. A
\texttt{StandardScaler} is fitted on training embeddings and reused to
standardize all embeddings; prototypes are defined in this scaled space, where each feature is centered and scaled to unit variance.

Using domain knowledge, we defined semantic prototype types grouped into non-defect and defect categories. The number of prototype types per class is itself a design hyperparameter of the method: too few prototypes cannot capture the visual diversity of each class, while too many dilute the semantics and increase the risk of redundant or ``dead'' prototypes.

Our initial design used only two non-defect prototypes (\emph{air}, \emph{healthy matrix}) and three defect prototypes (\emph{pores}, \emph{line-like defects}, \emph{mixed pores and lines}), with six expert-selected training patches per type. This configuration turned out to be problematic on the non-defect side. Inspecting the backbone embedding (see Fig.~\ref{fig:umap}), we observed that matrix patches containing part boundaries (edge patches, where matrix meets air) were actually \emph{further away} from pure matrix patches than pure air patches were from pure matrix. As a consequence, the two negative prototypes initialized from \emph{air} and \emph{healthy matrix} tended to coalesce: both ended up representing an ``uniform intensity'' concept (whether low or high), while all edge-containing matrix+air patches were pushed into the defect side of the latent space, where they did not belong. In other words, the latent geometry of the backbone was organized along a low-contrast vs.\ high-contrast axis rather than along a semantic defect vs.\ non-defect axis, which the two-prototype design could not accommodate.

\begin{figure}[H]
    \centering
    \includegraphics[width=1\linewidth]{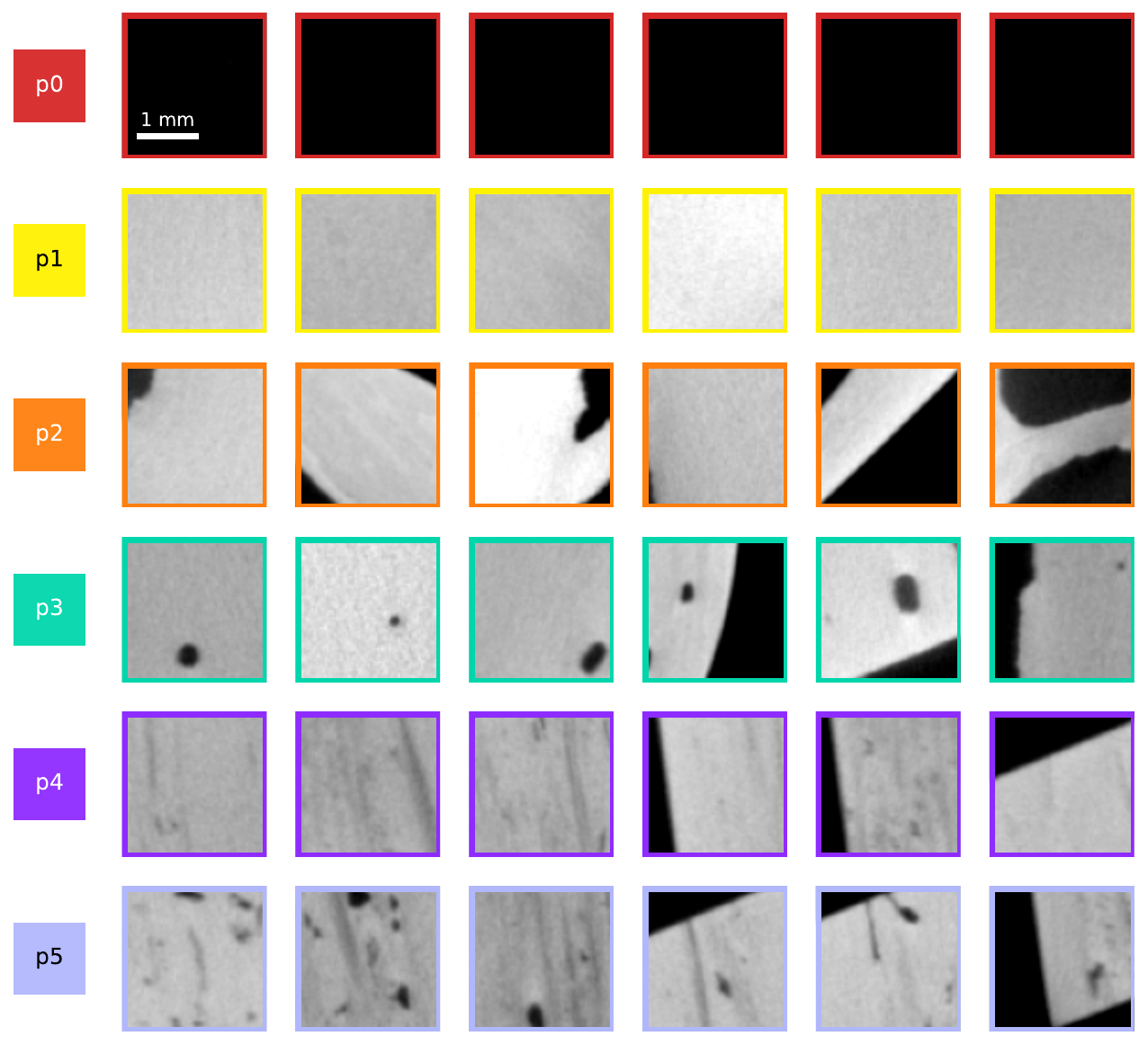}
    \caption{Labelled (expert) anchors for each prototype type. Grayscale intensities are
    windowed to $v_{\min}=0.2\times 256$ and $v_{\max}=0.8\times 256$.}
    \label{fig:pos_protos}
\end{figure}

To resolve this, we refined the taxonomy to three non-defect types (\emph{pure air}, \emph{healthy matrix}, \emph{matrix with air interface}) and kept the three defect types (\emph{pores}, \emph{line-like defects}, \emph{mixed pores and lines}), for a total of six semantic prototypes. The explicit \emph{matrix with air interface} prototype gives high-contrast non-defect patches a natural home in the latent space, and prevents edge-dominated but defect-free regions from being systematically misinterpreted as defects. For each type, an expert selected six representative training patches from the training set (36 patches in total), which we refer to as \emph{expert anchors} (Fig.~\ref{fig:pos_protos}). For the three defect types, the expert anchor selection was balanced so that three anchors contained part boundaries (edges) and three were fully internal, so that the learned prototypes do not conflate the presence of an edge with the presence of a defect. This yields a semi-supervised initialization in which prototype semantics are explicitly tied to expert-chosen examples. 

For clarity, in what follows we distinguish between \emph{expert anchors} (manually selected patches), \emph{prototypes} (learned latent vectors initialized from the medoid of the corresponding expert anchors), and \emph{learned anchors} (training patches that lie closest to a given prototype in the latent space).

\clearpage

In the scaled embedding space, the empirical centroid of the six expert
anchors for each type is used to initialize the corresponding prototype
vector. The prototype matrix $P$ thus contains one prototype per
semantic type and these prototypes are subsequently updated during training.

\begin{figure}[ht]
    \centering
    \tikzset{
      stage/.style={
        draw,
        rounded corners=3pt,
        minimum width=3.5cm,
        minimum height=1.05cm,
        align=center,
        font=\footnotesize
      },
      arrow/.style={-Latex, thick}
    }
    
    \begin{tikzpicture}[node distance=0.55cm, >=Latex]
    
    \node[font=\bfseries] (titleA) {(a) ResNet-50};
    \node[stage, below=0.65cm of titleA] (inA) {input\\$64\times64\times1$};
    \node[stage, fill=green!20, below=0.65cm of inA] (convA) {$7\times7$ conv, 64, /2};
    
    \node[stage, fill=green!20, below=0.35cm of convA] (c2A) {%
    $1\times1$ conv, 64\\
    $3\times3$ conv, 64\\
    $1\times1$ conv, 256};
    \node[font=\scriptsize, right=0.10cm of c2A] {$\times 3$};
    
    \node[stage, fill=green!20, below=0.35cm of c2A] (c3A) {%
    $1\times1$ conv, 128 /2\\
    $3\times3$ conv, 128\\
    $1\times1$ conv, 512};
    \node[font=\scriptsize, right=0.10cm of c3A] {$\times 4$};
    
    \node[stage, fill=green!20, below=0.35cm of c3A] (c4A) {%
    $1\times1$ conv, 256 /2\\
    $3\times3$ conv, 256\\
    $1\times1$ conv, 1024};
    \node[font=\scriptsize, right=0.10cm of c4A] {$\times 6$};
    
    \node[stage, fill=green!20, below=0.35cm of c4A] (c5A) {%
    $1\times1$ conv, 512 /2\\
    $3\times3$ conv, 512\\
    $1\times1$ conv, 2048};
    \node[font=\scriptsize, right=0.10cm of c5A] {$\times 3$};
    
    \node[stage, below=0.35cm of c5A] (fcA) {FC, 1};
    \node[stage, fill=yellow!30, below=0.65cm of fcA] (probA) {defect / non-defect};
    
    \draw[arrow] (inA) -- (convA);
    \draw[arrow] (convA) -- node[right=2pt, font=\scriptsize]{max pool, /2} (c2A);
    \draw[arrow] (c2A) -- (c3A);
    \draw[arrow] (c3A) -- (c4A);
    \draw[arrow] (c4A) -- (c5A);
    \draw[arrow] (c5A) -- node[right=2pt, font=\scriptsize]{avg pool} (fcA);
    \draw[arrow] (fcA) -- node[right=2pt, font=\scriptsize]{sigmoid + threshold} (probA);
    
    \node[font=\bfseries, right=3cm of titleA] (titleB) {(b) p-ResNet-50};
    \node[stage, below=0.65cm of titleB] (inB) {input\\$64\times64\times1$};
    \node[stage, fill=green!20, below=0.65cm of inB] (convB) {$7\times7$ conv, 64, /2};
    
    \node[stage, fill=green!20, below=0.35cm of convB] (c2B) {%
    $1\times1$ conv, 64\\
    $3\times3$ conv, 64\\
    $1\times1$ conv, 256};
    \node[font=\scriptsize, right=0.10cm of c2B] {$\times 3$};
    
    \node[stage, fill=green!20, below=0.35cm of c2B] (c3B) {%
    $1\times1$ conv, 128 /2\\
    $3\times3$ conv, 128\\
    $1\times1$ conv, 512};
    \node[font=\scriptsize, right=0.10cm of c3B] {$\times 4$};
    
    \node[stage, fill=green!20, below=0.35cm of c3B] (c4B) {%
    $1\times1$ conv, 256 /2\\
    $3\times3$ conv, 256\\
    $1\times1$ conv, 1024};
    \node[font=\scriptsize, right=0.10cm of c4B] {$\times 6$};
    
    \node[stage, fill=green!20, below=0.35cm of c4B] (c5B) {%
    $1\times1$ conv, 512 /2\\
    $3\times3$ conv, 512\\
    $1\times1$ conv, 2048};
    \node[font=\scriptsize, right=0.10cm of c5B] {$\times 3$};
    
    \node[stage, below=0.35cm of c5B] (headDesc) {prototype head};
    
    \node[stage, below=0.35cm of headDesc] (protoProb) {prototype probabilities};
    
    \node[stage, fill=yellow!30, below=0.55cm of protoProb] (clsB) {defect / non-defect};
    
    \draw[arrow] (inB) -- (convB);
    \draw[arrow] (convB) -- node[right=2pt, font=\scriptsize]{max pool, /2} (c2B);
    \draw[arrow] (c2B) -- (c3B);
    \draw[arrow] (c3B) -- (c4B);
    \draw[arrow] (c4B) -- (c5B);
    \draw[arrow] (c5B) -- node[right=2pt, font=\scriptsize]{avg pool} (headDesc);
    \draw[arrow] (headDesc) -- node[right=2pt, font=\scriptsize]{softmax} (protoProb);
    \draw[arrow] (protoProb) -- node[right=2pt, font=\scriptsize]{sum + threshold} (clsB);
    
    \end{tikzpicture}

    \caption{Network architectures. (a) Baseline ResNet-50 (b) Prototype-based variant (p-ResNet-50) that reuses the same grayscale ResNet-50 backbone (green), but replaces the FC head with a prototype head.}
    \label{fig:architecture}
\end{figure}
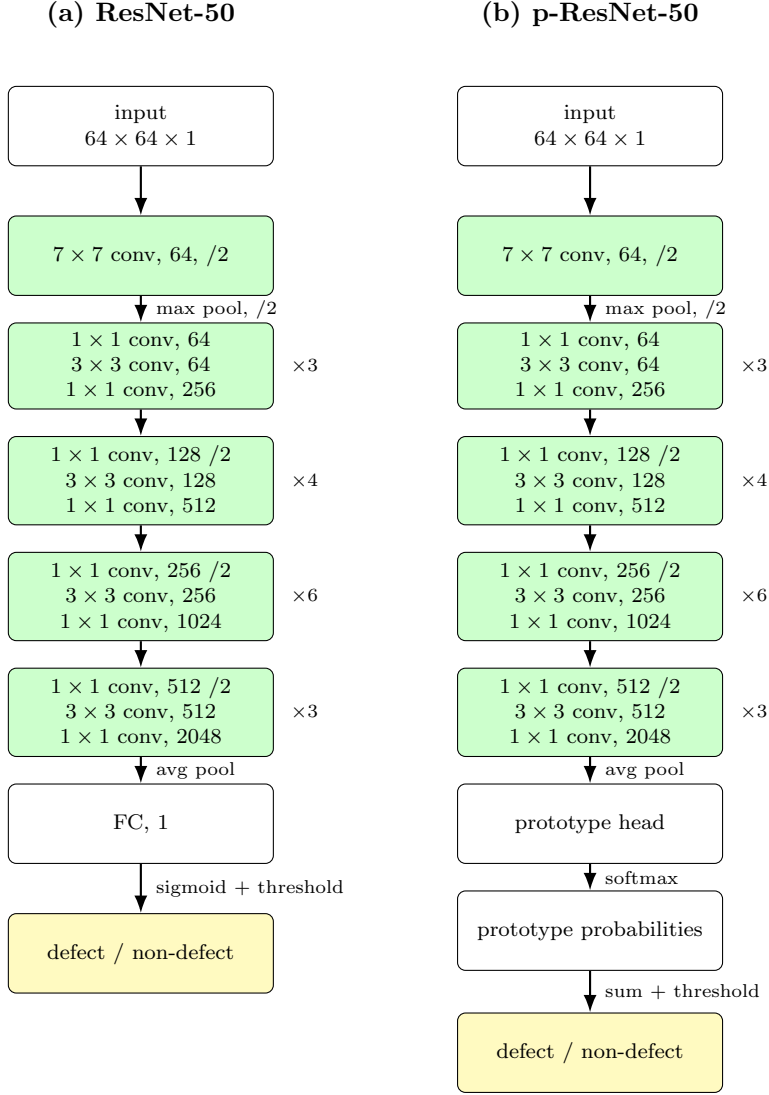

At inference time, the prototype head standardizes the input embedding $z(x)$ and computes the squared Euclidean distance to each prototype $P_k$, normalized by the embedding dimension. Distances are mapped to prototype logits via $-d^2 / \tau$. Class logits are obtained by log-sum-exp pooling of the prototype logits within each class, and class probabilities follow from a softmax over the two class logits. A global softmax over all prototype logits yields a distribution over the six prototypes, indicating which semantic type contributed most to the decision. The architectures of ResNet-50 and p--ResNet-50 are summarized
in Fig.~\ref{fig:architecture}.

\subsection*{p--ResNet-50 training}

Training of p--ResNet-50 uses a composite loss that combines classification accuracy with constraints on the prototype geometry, as shown in Equation~\ref{eq:proto-loss}. 

The prototype head comprises $k_{\text{neg}} = 3$ negative and $k_{\text{pos}} = 3$ positive prototypes (6 in total). Initial experiments used centroid-based initialization, in which each prototype was set to the mean of the expert candidate embeddings for its subtype. This scheme proved problematic in practice: the centroid is a synthetic point that does not correspond to any real training patch, cannot be directly inspected or visualized as an exemplar, and, when the candidate set spans a slightly heterogeneous region of the latent space, lands in a low-density area between real samples, weakening the semantic grounding of the prototype and making the subsequent $\mathcal{L}_{\text{anchor}}$ and $\mathcal{L}_{\text{medoid}}$ terms less meaningful. We therefore switched to a medoid-based initialization, in which each prototype is set to the expert candidate whose embedding is closest to the centroid of its subtype. This guarantees that every prototype corresponds to a real, inspectable training sample at initialization, while still sitting at the geometric core of its expert-curated candidate set.

$\mathcal{L}_{\text{cls}}$ is the cross-entropy on the binary label. $\mathcal{L}_{\text{pull}}$ pulls embeddings towards same-class prototypes, and $\mathcal{L}_{\text{push}}$ pushes embeddings away from opposite-class prototypes using an exponential penalty with scale $\tau_{\text{push}}$. These three terms follow the standard formulation in prototype-based classification networks~\cite{chen2019_protopnet,nauta2021_prototree,rymarczyk2021_prototype}. $\mathcal{L}_{\text{div}}$ encourages diversity between $\ell_2$-normalized prototypes of the same class (margin $\delta$), and $\mathcal{L}_{\text{ent}}$ and $\mathcal{L}_{\text{usage}}$ promote a moderately diffuse within-class assignment to prototypes and discourage ``dead'' prototypes that are never selected. These terms adapt diversity and usage regularizers proposed for ProtoPNet-style models by (i) restricting the diversity penalty to prototypes of the same class and enforcing a margin on their pairwise cosine similarities, in order to avoid within-class prototype collapse while allowing classes to remain well separated, and (ii) computing entropy and usage penalties from the normalized prototype-assignment distribution produced by our head, rather than from fixed assignment heuristics, so that regularization acts directly on the probabilistic prototype attributions used at inference.

$\mathcal{L}_{\text{anchor}}$ uses the expert anchors\footnote{Beyond enforcing semantic continuity between the expert-defined anchors and the learned prototypes, this term proved crucial in practice to prevent prototype collapse: in our experiments, adding $\mathcal{L}_{\text{anchor}}$ was the main modification that consistently stabilized training and encouraged prototypes to converge toward distinct, semantically meaningful regions of the latent space.} to keep each prototype close to its designated semantic type, thereby explicitly tethering the latent representation to domain-defined defect categories. We further introduce an $\mathcal{L}_{\text{medoid}}$ term (added to Equation~\ref{eq:proto-loss} as $+\,\lambda_{\text{medoid}}\,\mathcal{L}_{\text{medoid}}$), which penalizes the squared deviation of each learned prototype from its initialization medoid position in the standardized embedding space. This anchoring to the original medoid locations stabilizes the prototype geometry during the early optimization phase, when the classification and pull/push gradients can otherwise drag prototypes away from their semantically curated initializations; accordingly. $\mathcal{L}_{\text{medoid}}$ is active throughout training, as the backbone (layers 1–3) remains frozen for the full run. The last two terms regularize the prototype norms and the temperature~$\tau$, preventing unbounded growth of prototype vectors and degenerate, overly sharp or overly flat similarity distributions. 

The medoid- and mean-based prototype initializations are standard baselines in the prototype-network literature~\cite{snell2017_prototypical}. What is, to the best of our knowledge, specific to the present work is rather the way these ingredients are combined: medoids computed over small, expert-curated candidate sets for each defect subtype (rather than over full class populations or unsupervised clusters). On top of this initialization, we add two explicit regularization terms $\mathcal{L}_{\text{anchor}}$ and $\mathcal{L}_{\text{medoid}}$  —the latter restricted to the frozen-backbone phase — and we also include a temperature penalty of the form $\lambda_{\tau}\lvert \tau - 1 \rvert$.

\begin{equation}
\begin{split}
\mathcal{L}
= {} & \lambda_{\text{cls}} \, \mathcal{L}_{\text{cls}}
    + \lambda_{\text{pull}} \, \mathcal{L}_{\text{pull}}
    + \lambda_{\text{push}} \, \mathcal{L}_{\text{push}} \\
  & + \lambda_{\text{div}} \, \mathcal{L}_{\text{div}}
    + \lambda_{\text{ent}} \, \mathcal{L}_{\text{ent}}
    + \lambda_{\text{usage}} \, \mathcal{L}_{\text{usage}} \\
  & + \lambda_{\text{anchor}} \, \mathcal{L}_{\text{anchor}}
    + \lambda_{\text{medoid}} \, \mathcal{L}_{\text{medoid}} \\
  & + \lambda_{\text{proto}} \, \lVert P \rVert_2^2
    + \lambda_{\tau} \, \lvert \tau - 1 \rvert .
\end{split}
\label{eq:proto-loss}
\end{equation}
The hyperparameters were fixed to
\(
(\lambda_{\text{cls}},\allowbreak \lambda_{\text{pull}},\allowbreak \lambda_{\text{push}},\allowbreak
 \lambda_{\text{div}},\allowbreak \lambda_{\text{ent}},\allowbreak \lambda_{\text{usage}},\allowbreak
 \lambda_{\text{anchor}},\allowbreak \lambda_{\text{medoid}},\allowbreak \lambda_{\text{proto}},\allowbreak \lambda_{\tau},\allowbreak
 \tau_{\text{push}},\allowbreak \delta,\allowbreak \tau)
= (0.1,\allowbreak 0.05,\allowbreak 0.01,\allowbreak 1.0,\allowbreak 0.01,\allowbreak 0.1,\allowbreak
   2.0,\allowbreak 0.5,\allowbreak 10^{-6},\allowbreak 10^{-4},\allowbreak 0.1,\allowbreak 0.7,\allowbreak 1)
\).
These values were set heuristically, guided by preliminary runs rather than an exhaustive search. Our guiding objective was twofold: to preserve the backbone's classification performance, and to enforce semantic continuity between the expert-defined anchor categories and the learned prototypes. A systematic sensitivity analysis of this configuration is left for future work.

The p--ResNet-50 network is initialized from the supervised ResNet-50
checkpoint used in the baseline comparison. The prototype head is
attached to this backbone, and training fine-tunes the prototype head
together with the last ResNet block (layer4). Earlier layers are frozen.

Optimization uses AdamW with learning rates of $5\times10^{-4}$ for the
prototype head and $5\times10^{-6}$ for the trainable backbone parameters
(weight decay $10^{-4}$). Gradients on the prototype head and layer4
are clipped to a maximum $\ell_2$-norm of 1.0. Class imbalance is
handled via inverse-frequency class weights in $\mathcal{L}_{\text{cls}}$.
The same data augmentation strategy as for the baseline was employed.

A \texttt{ReduceLROnPlateau} scheduler monitors the validation value of
the composite loss~\eqref{eq:proto-loss} and reduces both learning rates
by a factor of 0.5 after 10 epochs without improvement (minimum
$10^{-7}$). Early stopping is triggered if the validation loss does not
improve for 50 consecutive epochs, and the checkpoint with the lowest
validation loss is used for evaluation. As for the baseline, the final
decision threshold is selected on the validation set by maximizing the
F\textsubscript{1} score.

\section{Results}
The different ResNet backbones were systematically compared under the same training and evaluation protocol. All architectures achieved near-ceiling discriminative performance on the held-out test set, with ROC--AUC values $\geq 0.991$ and PR--AUC values $\geq 0.986$ (Table~\ref{tab:resnet_perf}). Because of the test set's moderate size, small changes in the number of misclassified patches can lead to noticeable shifts in the point estimates, thus explaining the observed non-monotonic behavior. The test-set expected calibration errors remain low for all models after temperature scaling, suggesting that reusing the validation set for calibration did not lead to noticeable overfitting of the calibration step.

95\% confidence intervals (CIs) were therefore estimated via bootstrap resampling at the patch level and are reported in Table~\ref{tab:resnet_perf}. The CIs for the different backbones show substantial overlap, indicating that deeper architectures do not yield large systematic gains in this data regime; however, ResNet-50 consistently exhibits the most favorable combination of accuracy, F1-score, specificity and calibration. We therefore selected ResNet-50 as the backbone for all subsequent experiments.

\begin{table}[H]
\centering
\caption{Patch-level performance of the ResNet classifiers on the test set. Metrics are reported at the validation-selected operating threshold for each model. Values in parentheses denote 95\% bootstrap confidence intervals (patch level).}
\label{tab:resnet_perf}
\resizebox{\textwidth}{!}{%
\begin{tabular}{lccccc}
\hline
Metric & ResNet-18 & ResNet-34 & ResNet-50 & ResNet-101 & ResNet-152 \\
\hline
Threshold &
0.371 &
0.361 &
0.361 &
0.381 &
0.331 \\
Accuracy &
\makecell{0.952 \\ (0.933--0.971)} &
\makecell{0.945 \\ (0.921--0.964)} &
\makecell{0.959 \\ (0.940--0.978)} &
\makecell{0.947 \\ (0.923--0.967)} &
\makecell{0.950 \\ (0.931--0.969)} \\
Precision &
\makecell{0.907 \\ (0.862--0.952)} &
\makecell{0.933 \\ (0.889--0.972)} &
\makecell{0.925 \\ (0.884--0.964)} &
\makecell{0.896 \\ (0.849--0.937)} &
\makecell{0.907 \\ (0.863--0.949)} \\
Rec/Sens &
\makecell{0.967 \\ (0.934--0.993)} &
\makecell{0.914 \\ (0.866--0.955)} &
\makecell{0.967 \\ (0.936--0.993)} &
\makecell{0.967 \\ (0.935--0.993)} &
\makecell{0.961 \\ (0.927--0.987)} \\
F1 &
\makecell{0.936 \\ (0.906--0.963)} &
\makecell{0.924 \\ (0.890--0.951)} &
\makecell{0.945 \\ (0.918--0.970)} &
\makecell{0.930 \\ (0.899--0.956)} &
\makecell{0.933 \\ (0.904--0.959)} \\
Specificity &
\makecell{0.944 \\ (0.915--0.972)} &
\makecell{0.962 \\ (0.938--0.985)} &
\makecell{0.955 \\ (0.929--0.978)} &
\makecell{0.936 \\ (0.906--0.961)} &
\makecell{0.944 \\ (0.917--0.969)} \\
ROC--AUC &
0.992 & 0.986 & 0.993 & 0.989 & 0.991 \\
PR--AUC &
0.988 & 0.981 & 0.990 & 0.985 & 0.984 \\
ECE &
0.038 & 0.047 & 0.024 & 0.025 & 0.046 \\
Brier &
0.033 & 0.040 & 0.029 & 0.035 & 0.036 \\
Confusion (TP, FP, TN, FN) &
(147, 15, 251, 5) &
(139, 10, 256, 13) &
(147, 12, 254, 5) &
(147, 17, 249, 5) &
(146, 15, 251, 6) \\
\hline
\end{tabular}%
}
\end{table}

At its validation-selected operating point (threshold $=0.361$), ResNet-50 achieved an accuracy of $0.959$ on the test set, with a recall (sensitivity) of $0.967$ and a precision of $0.925$, corresponding to an F1-score of $0.945$ and a specificity of $0.955$ (confusion counts: $\text{TP}=147$, $\text{FP}=12$, $\text{TN}=254$, $\text{FN}=5$). Post-hoc temperature scaling with a learned temperature $T = 1.612$ yielded well-calibrated probabilities, as reflected by an expected calibration error of $0.024$ and a Brier score of $0.029$.

Using the same training considerations, the p-ResNet-50 was trained and evaluated. The temperature scaling was again fitted on the validation set ($T = 0.608$) and, at the validation-selected operating point (threshold $= 0.511$), the model attains a ROC--AUC of $0.994$ and a PR--AUC of $0.991$. At this operating point, recall (sensitivity) is $0.914$ and precision is $0.965$, yielding an F1-score of $0.939$, with confusion counts $\text{TP} = 139$, $\text{FP} = 5$, $\text{TN} = 261$, and $\text{FN} = 13$. Calibration assessed after temperature scaling results in an expected calibration error (ECE) of $0.027$ and a Brier score of $0.029$. As above, 95\% confidence intervals (CIs) were estimated via bootstrap resampling at the patch level. The complete set of patch-level results is reported in Table~\ref{tab:pnet_perf}.

\begin{table}[H]
\centering
\caption{Patch-level performance of the prototype-based ResNet-50 (p--ResNet) on the test set. Metrics are reported at the validation-selected operating threshold (0.511) after temperature scaling ($T = 0.608$). Values in parentheses denote 95\% bootstrap confidence intervals (patch level).}
\label{tab:pnet_perf}
\resizebox{\textwidth}{!}{%
\begin{tabular}{ccccccccc}
\hline
Accuracy & Precision & Rec/Sens & F1 & Specificity & ROC--AUC & PR--AUC & ECE & Brier \\
\hline
\makecell{0.957 \\ (0.935--0.976)} &
\makecell{0.965 \\ (0.934--0.993)} &
\makecell{0.914 \\ (0.868--0.959)} &
\makecell{0.939 \\ (0.909--0.966)} &
\makecell{0.981 \\ (0.964--0.996)} &
0.994 &
0.991 &
0.027 &
0.029 \\
\hline
\end{tabular}%
}
\\[0.4em]
\small \textit{Confusion counts:} TP = 139, FP = 5, TN = 261, FN = 13.
\end{table}

From a purely discriminative point of view, the prototype-based ResNet-50 performs comparably to the standard (black-box) ResNet-50, but does not clearly dominate it. At their respective validation-selected operating points, the black-box model attains slightly higher accuracy and F1-score (accuracy $0.959$ vs.\ $0.957$, F1 $0.945$ vs.\ $0.939$), primarily due to its markedly higher sensitivity (recall $0.959$ vs.\ $0.914$). In contrast, the prototype-based network trades some recall for improved precision and specificity (precision $0.925$ vs.\ $0.965$, specificity $0.955$ vs.\ $0.981$), thus reducing the number of false positives. The ROC--AUC and PR--AUC of p--ResNet-50 are very close to those of the black-box ResNet-50 (Table~\ref{tab:resnet_perf}, Table~\ref{tab:pnet_perf}), with comparable Brier scores, whereas its ECE after calibration is slightly higher. Importantly, the bootstrap confidence intervals for all metrics show substantial overlap, indicating that, given the test-set size, the observed differences are compatible with a regime in which the two models offer broadly similar discriminative performance, with a modest but interpretable trade-off between sensitivity and precision.

This modest performance gap can be understood as an instance of the well-known accuracy--interpretability trade-off, sometimes informally referred to as an ``interpretability tax'': constraining a model to reason through a small set of semantically meaningful prototypes, anchored to expert-defined defect categories, restricts the hypothesis space compared to an unconstrained deep classifier, and this restriction can translate into a small loss of raw predictive performance~\cite{rudin2019_stop_explaining}. In our setting, this tax takes the specific form of a slight drop in sensitivity and F1-score relative to the black-box ResNet-50, while ROC--AUC, PR--AUC and specificity remain essentially on par. We argue that, in a non-destructive testing context where each positive prediction must ultimately be reviewed and acted upon by a human expert, paying such a modest tax is a reasonable price for a model whose decisions can be directly traced back to expert-curated prototypes, and whose errors can be inspected, audited, and understood at the level of individual defect categories rather than through post-hoc saliency maps on an opaque backbone.

\begin{figure}[h]
    \centering
    \includegraphics[width=1\linewidth]{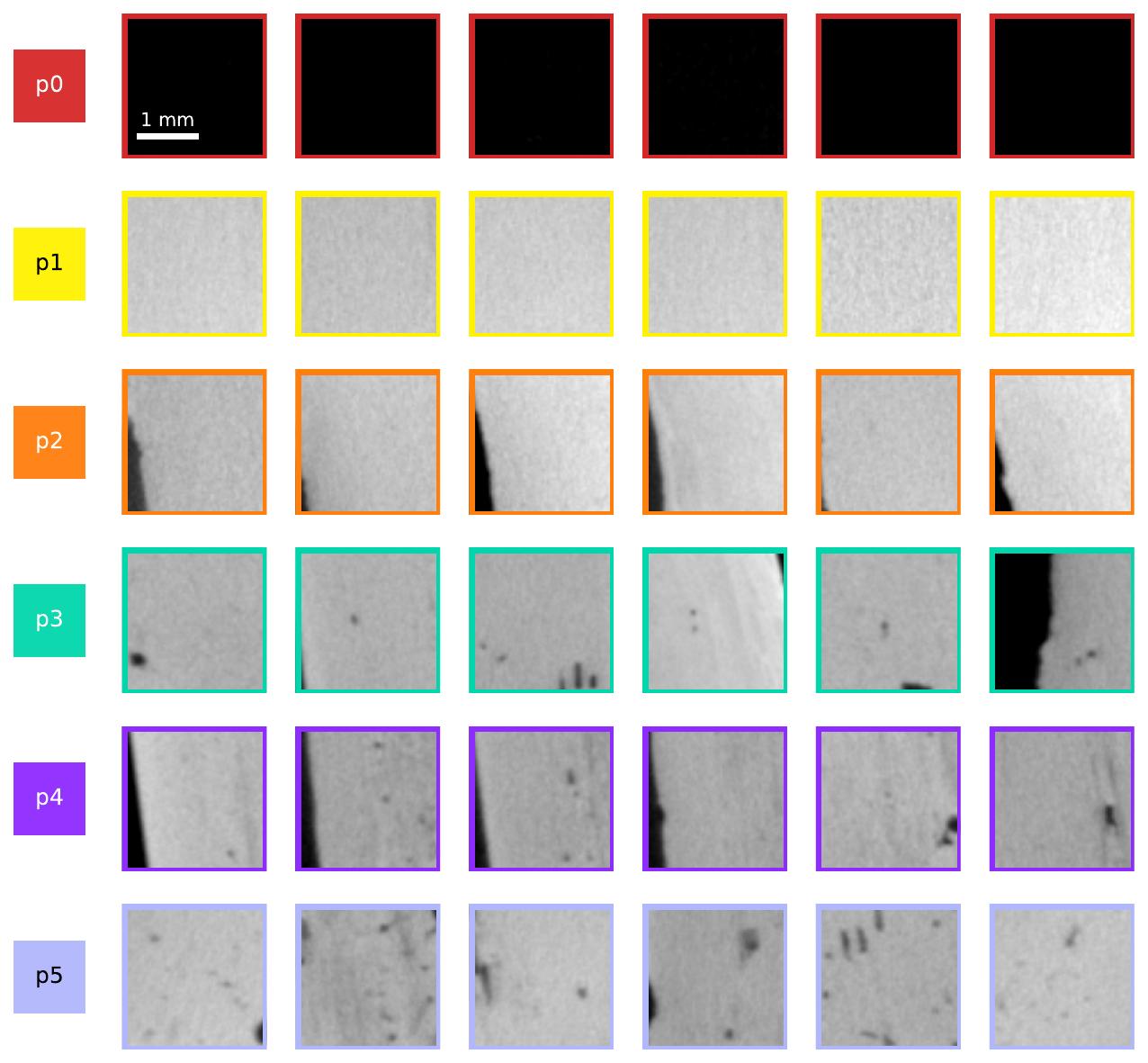}
    \caption{Nearest (learned) anchors for the six semantic prototype types.
Each row corresponds to one prototype ($p0$–$p5$), and all patches shown
are training samples whose embeddings lie closest to that prototype in
the latent space (i.e., \emph{learned anchors}, distinct from the expert
anchors of Fig.~\ref{fig:pos_protos}). Within each row, the leftmost
patch is the closest anchor, followed by additional close neighbours.
Border colors are shared within each row and match the semantic
categories defined in Fig.~\ref{fig:pos_protos}. Grayscale intensities
are windowed to $v_{\min}=0.2\times 256$ and $v_{\max}=0.8\times 256$.}
    \label{fig:anchors}
\end{figure}

Fig.~\ref{fig:anchors} shows, for each prototype, the training patches whose embeddings lie closest to it in the latent space; we refer to these nearest neighbours as \emph{learned anchors}. Consistent with the semi-supervised initialization based on expert anchors (Fig.~\ref{fig:pos_protos}), the learned anchors for each prototype remain aligned with the six semantic categories (\emph{non-defect}: \emph{pure air}, \emph{air with pores}, \emph{healthy matrix}; \emph{defect}: \emph{pores}, \emph{line-like defects}, \emph{mixed pores and lines}), confirming that the intended domain-informed structure is preserved. 

It is worth noting that, despite the global image normalization applied at the preprocessing stage---which rescales all pixel intensities to a common range and could in principle wash out absolute density differences---the model still cleanly separates the \emph{pure air} and \emph{healthy matrix} non-defect prototypes, indicating that the backbone has learned to encode local texture and contextual cues beyond mean intensity alone. Semantic continuity is therefore broadly preserved across all six categories; some residual class intermingling is nevertheless observed among the positive (defect) prototypes, most noticeably in the form of a few isolated pores appearing within patches assigned to the \emph{line-like defects} prototype ($p4$). 

This behavior is consistent with the fact that real defect patches rarely contain a single defect type in isolation, and the prototype attribution reflects the \emph{dominant} semantic content of the patch rather than a strict partition. In general, the expert-selected anchors do not themselves become prototypes, nor do they usually coincide with the closest learned anchors: during training, the model converges toward different but visually similar patches that it deems more representative under the learned similarity metric. 

Further work could explore additional mechanisms---such as finer-grained expert annotations, stricter anchor-side constraints, or prototype-specific purity penalties---to further sharpen semantic continuity and reduce residual mixing between closely related defect categories.

\clearpage

\begin{figure}[h]
    \centering
    \includegraphics[width=1.0\linewidth]{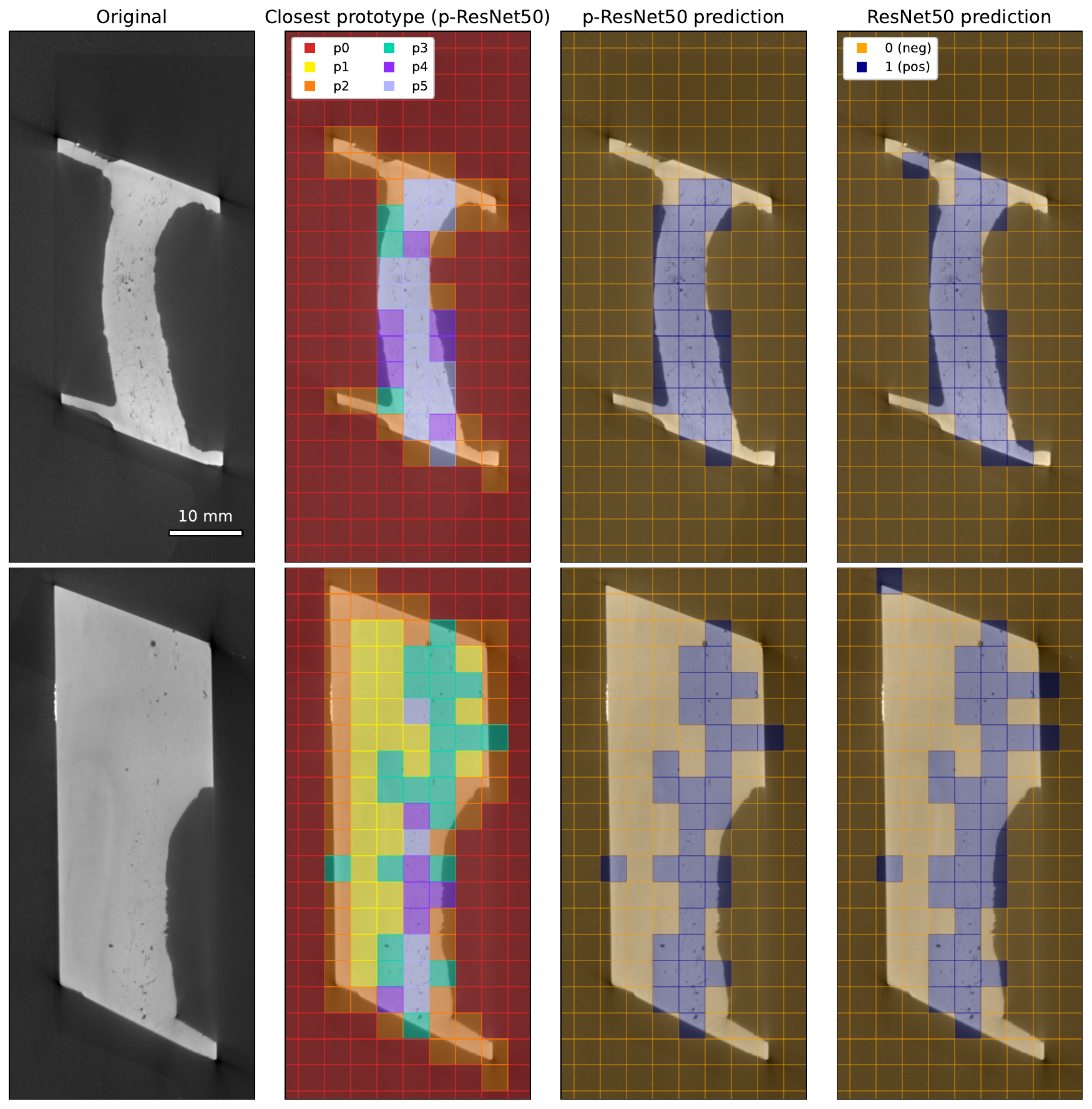}
    \caption{Patch-wise comparison of the prototype-based network and the
black-box ResNet-50 on two tomographic sections. The first column shows
the original reconstructions (with the patch grid overlaid). The second
column indicates, for each patch, the index of the closest prototype
learned by p--ResNet-50 (color-coded as in Fig.~\ref{fig:anchors}).
The third and fourth columns display the corresponding binary defect
predictions of p--ResNet-50 and ResNet-50, respectively.}
    \label{fig:comparison}
\end{figure}

Fig.~\ref{fig:comparison} shows two representative tomographic sections together with patch-wise prototype assignments and predictions. As illustrated, the two models produce very similar defect maps and agree on the global classification, with only minor local discrepancies near boundaries and in low-contrast regions. For the p--ResNet-50, the prototype attribution is also shown, enabling a semantic interpretation of each patch in terms of the previously defined categories.

Beyond reproducing the black-box predictions, this visualization illustrates that the prototype-based model not only indicates whether a patch is defective, but also provides an explicit rationale for its decision: a patch is classified as defective because its appearance is most similar to a learned prototype corresponding to, for instance, a pore or a line-like defect. Patch-level maps can be further refined by using a smaller stride and aggregating overlapping predictions via majority voting, yielding smoother and more precise defect delineations. These dense predictions, in turn, enable quantitative analyses such as estimating defect yield at the component level and per defect type, or characterizing the spatial distribution of defects across predefined zones of the part. By defining application-specific thresholds on these quantities, or by identifying critical regions and defect types to avoid, one can derive operational metrics that NDT experts may use to accept or reject components in a principled and reproducible manner.

\begin{figure}[H]
    \centering
    \includegraphics[width=1.0\linewidth]{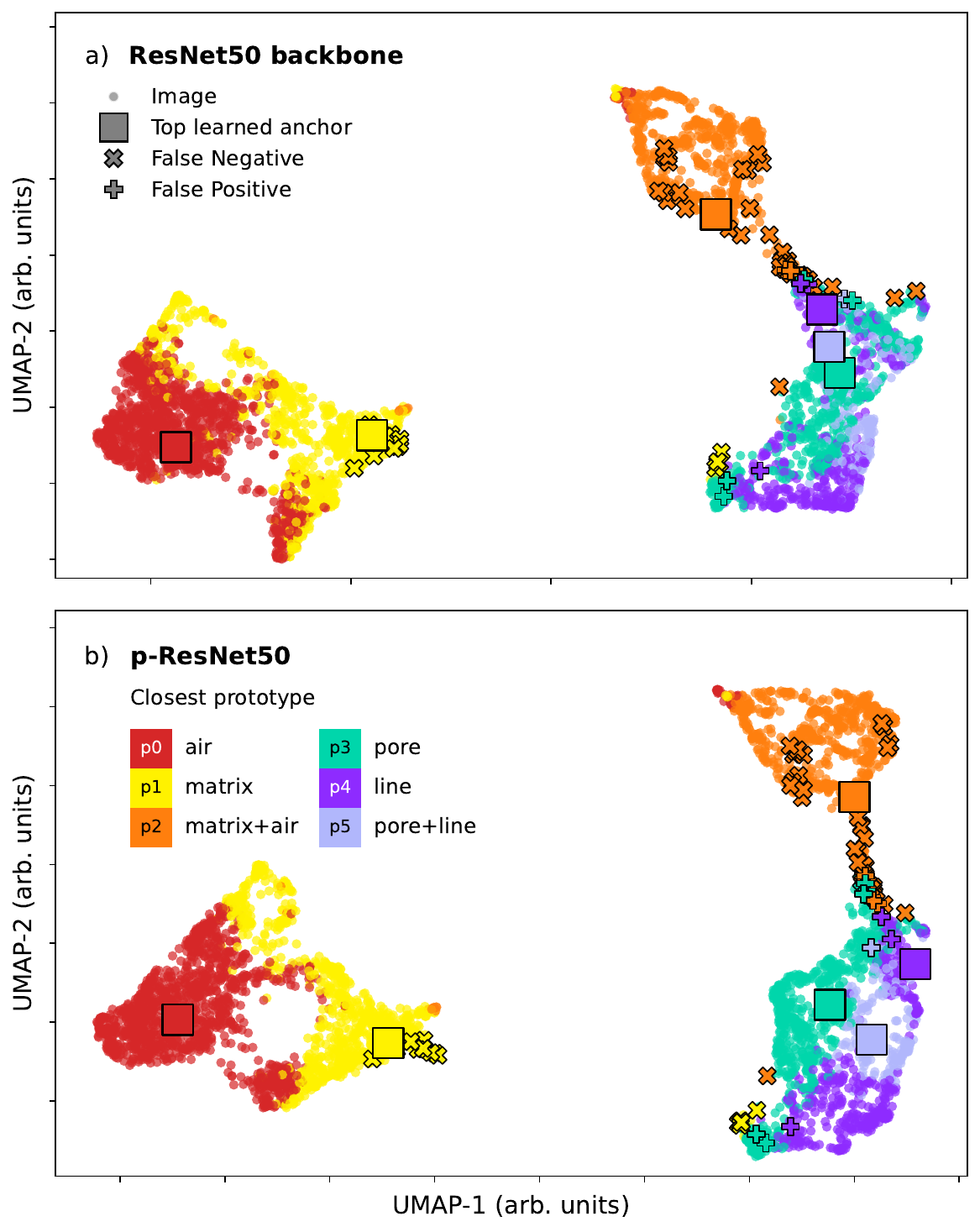}
    \caption{UMAP projection of patch-level embeddings for the ResNet-50
(a) and p--ResNet-50 (b) models. Each point corresponds to a
patch from the train/validation/test sets. Colors indicate the index of
the closest prototype in the p--ResNet-50 latent space (p0: air, p1:
matrix, p2: matrix+air, p3: pores, p4: line-like defects, p5: mixed
pores and lines); square markers denote the expert anchors. False
negatives and false positives on the test set are highlighted with
cross and plus symbols, respectively. The prototype-based model
organizes the latent space into compact, semantically coherent regions
while localizing zones of uncertainty where most misclassifications
occur.}
    \label{fig:umap}
\end{figure}

To further analyze the representations learned by both architectures, we projected the patch-level embeddings onto two dimensions using UMAP, using all patches from the training, validation and test sets (Fig.~\ref{fig:umap}). a) shows the embedding of the standard ResNet-50 and b) the embedding of the p--ResNet-50. In both cases, each point corresponds to a patch, and the color encodes the index of the closest prototype in the p--ResNet-50 latent space.\footnote{In both panels, colors are assigned according to the index of the closest prototype of the p--ResNet-50 in its latent space. For the ResNet-50 embedding, the p--ResNet-50 prototypes are projected and reused solely for coloring, even though ResNet-50 does not have its own prototype head. This allows a direct comparison of the semantic attribution structure between the two models.} The square markers represent the top learnt anchors for each prototype class.

At a coarse level, both embeddings exhibit the same global topology: the latent space splits into two clearly disconnected ``islands'' that correspond to the underlying image content rather than to the defect/non-defect dichotomy. The first island groups all low-contrast patches, i.e.\ patches whose pixel values are either uniformly high (pure matrix, p1) or uniformly low (pure air, p0); by construction, these patches cannot contain defects, and the whole island is therefore populated exclusively by non-defect prototypes. The second island groups all high-contrast patches, which contain strong intensity variations either because of air/matrix interfaces (p2: matrix+air, still a non-defect category) and because of characteristic defect motif and/or edges (p3: pores, p4: line-like defects). This second island is the one in which prototype assignments can legitimately span both non-defect (p2) and defect (p3, p4, p5) categories, and it therefore constitutes the ``uncertainty island'' of the model: it is the only region of the latent space where the binary decision is not trivially dictated by the image content. This two-island topology is already visible in the ResNet-50 backbone embedding (Fig.~\ref{fig:umap}a), indicating that the separation between low- and high-contrast patches is primarily learned during the standard classification pre-training, and is not an artifact of the prototype head.

The effect of prototype training is instead to \emph{restructure} the interior of each island. In the low-contrast island, the black-box ResNet-50 already separates pure air (p0) from pure matrix (p1) to some extent, but the two regions remain loosely organized; after prototype training, the p--ResNet-50 embedding (Fig.~\ref{fig:umap}b) shows a visibly cleaner split between p0 and p1, with each expert anchor sitting at the core of its respective region. This is a non-trivial observation: since the preprocessing pipeline normalizes all pixel intensities to a common range, absolute density information is partly washed out, and the fact that the model nonetheless maintains a stable separation between pure air and pure matrix patches indicates that the backbone has learned to exploit local texture and contextual cues beyond mean intensity alone. 

The effect is even more pronounced in the high-contrast island. In the ResNet-50 backbone, the different semantic categories (p2--p5) are almost completely intermixed, and the expert anchors of different classes sit close to one another in overlapping regions, making it essentially impossible to read off a semantic structure directly from the black-box embedding. This retrospectively justifies the need to retrain the network with an explicit prototype structure. After p--ResNet-50 training, coherent sub-domains emerge for each prototype: the anchors of different classes are clearly pulled apart and each come to anchor their own region. Notably, the mixed pores-and-lines prototype (p5) settles geometrically \emph{between} the pure pores (p3) and pure lines (p4) regions, in agreement with its hybrid semantic content, and both p3 and p4 remain connected to the matrix+air region (p2), consistent with the fact that isolated pores and short line defects share edge-like local structure with simple air/matrix interfaces.

Despite this reorganization, an ``uncertainty zone'' remains, and it is precisely where one would expect it: in the transition regions of the high-contrast island, where prototype attribution can switch between p2, and p3 or p4. Some classification errors are also found in the p1 zone. All misclassifications observed on the test set fall in these four categories, while no errors are observed for samples dominated by the pure air (p0) or the mixed pores-and-lines (p5) prototypes. 

Taken together, these observations indicate that the prototype-based network preserves the global two-island topology inherited from the black-box backbone while restructuring the interior of each island into interpretable, semantically coherent sub-domains. This embedding-level view offers a concrete tool for delimiting the \emph{domain of applicability} of the model: it makes explicit which defect types and spatial contexts are reliably recognized (e.g., mixed pores-and-lines, and more generally any patch falling deep inside a well-separated prototype region) and which configurations tend to produce ambiguity (transitions between matrix, matrix+air, pores and line-like defects in the high-contrast island). Such an analysis helps NDT experts identify the strengths and weaknesses of the system, audit its predictions at the level of individual prototypes, and decide under which acquisition and material conditions the model can be trusted to sanction or accept components.

\section{Conclusion}

In this work, we investigated interpretable deep learning for defect
detection in X-ray tomography of aerospace-grade SiC/SiC composites.
Starting from a curated patch-level dataset drawn from four defect-rich
specimens, we established a supervised baseline by systematically
evaluating a family of residual networks under a common training,
calibration and evaluation protocol. All backbones achieved
near-ceiling performance on the held-out test set, and ResNet-50
emerged as the most favorable trade-off between accuracy,
F\textsubscript{1}-score, specificity and calibration, providing a
solid black-box reference for subsequent experiments and a reusable
benchmark for future work on this dataset.

Building on this backbone, we designed a prototype-based extension that
preserves competitive discriminative performance while yielding
semantically grounded, case-based explanations. The prototype network
delivers patch-wise defect maps that closely match those of the
black-box model, with only modest reductions in sensitivity and
F\textsubscript{1}-score, but offers substantially richer outputs:
each decision can be traced back to a small set of learned prototypes
aligned with expert-defined semantic categories (air, healthy matrix,
pores, line-like defects, mixed pores and lines). On the algorithmic
side, we introduce a tailored training objective for prototype
networks that combines standard pull/push terms with adapted diversity
and usage regularizers and a novel anchor-based term, together with a
temperature regularizer, which jointly mitigate prototype collapse and
promote semantically meaningful prototype clusters aligned with expert
anchors.

Beyond raw accuracy, we assessed aspects of robustness through
held-out calibration metrics, bootstrap confidence intervals and
embedding-level analyses. UMAP visualizations showed that the prototype-based model organizes
embeddings into compact, semantically coherent regions with clearly
localized “uncertainty islands” in which most misclassifications
occur. This structure not only clarifies which defect types and
spatial contexts are reliably recognized, but also helps delineate the
domain of applicability of the model and identify borderline regimes
where expert oversight remains critical. 

Several limitations of the present study suggest directions for future
work. The experiments were conducted on a limited number of laboratory
specimens and focused on a binary defect/non-defect formulation with
2D patch-based inputs; extending the approach to larger, more diverse
industrial datasets, fully 3D architectures, and multi-class defect
taxonomies will be important steps toward deployment and a more
thorough assessment of robustness. In addition, the prototype loss
configuration was fixed heuristically, and a systematic study of its
hyperparameters, together with more advanced uncertainty
quantification and active-learning strategies, could further improve
both performance and interpretability. Nonetheless, the results
demonstrate that prototype-based architectures offer a promising
compromise between accuracy, robustness and transparency for XCT
inspection of SiC/SiC composites, and provide practical tools for
auditing and trustworthy integration of deep learning into industrial
NDT workflows.

\section*{Acknowledgments}
This work was conducted as part of the ``COSMIC'' project between IRT Saint Exup\'ery and
Safran Ceramics, with financial support from Safran Ceramics and the
French National Research Agency (ANR). The authors are grateful
to Franck Mamalet and Ahmad Berjaoui for their careful reading of the manuscript and
their insightful suggestions. The authors declare no conflict of interest.

\section*{Data availability}

All resources necessary to reproduce the experiments reported in this work are publicly available at https://huggingface.co/datasets/antpeacor/cosmicsicsic . This includes:

\begin{itemize}
    \item the curated patch dataset used for training, validation and testing, together with
    the corresponding train/validation/test splits and binary labels (defect / non-defect);
    \item the trained weights of the baseline ResNet-50 classifier and the prototype-based
    p--ResNet-50 model;
    \item the full model definitions (network architectures), prototype-initialization
    routines, and training scripts for both the baseline and prototype-based networks;
    \item the code used to generate the quantitative results (metrics, confidence intervals),
    prototype visualizations, patch-wise prediction maps and UMAP embeddings.
\end{itemize}

The raw XCT volumes of the SiC/SiC parts can be shared for research purposes upon
reasonable request to the corresponding author, subject to the approval of the
industrial partners.

\bibliographystyle{unsrt} 
\bibliography{references} 

\end{document}